\documentclass[letterpaper,10pt,conference]{ieeeconf}

\IEEEoverridecommandlockouts
\usepackage{amsmath}
\usepackage{amssymb}
\usepackage{graphicx}
\usepackage{multirow}
\usepackage{booktabs}
\usepackage{tabularx}

\title{\LARGE \bf
A Cyclic Adaptation-Generalization Framework with Uncertainty-Guided Self-Paced Learning for \\Long-Term Brain-Machine Interfaces
}

\author{Jiyu Wei$^{1,2}$, Di Hong$^{1,2}$, Zhanjie Zhang$^{1}$, Dazhong Rong$^{*,1,2}$, Qinming He$^{1,2}$, Yueming Wang$^{1,2}$
\\$^{1}$College of Computer Science and Technology, Zhejiang University, Hangzhou, China
\\$^{2}$Nanhu Brain-Computer Interface Institute, Hangzhou, China
\\{\tt\{weijiyu,hongd,cszzj,rdz98,hqm,ymingwang\}@zju.edu.cn}
\thanks{*Corresponding author: Dazhong Rong.}
}

\begin{document}

\maketitle
\thispagestyle{empty}
\pagestyle{empty}

\begin{abstract}
Brain-Machine Interfaces (BMIs), which link the brain to external devices, hold great potential in rehabilitation, human performance augmentation, and human-centered robotics. However, invasive BMIs face a critical challenge for long-term deployment due to neural drift, which degrades decoding performance over time and necessitates frequent recalibration. Existing methods designed to mitigate neural drift typically rely on either domain adaptation (DA) or domain generalization (DG) alone and often fail to capture fine-grained distribution shifts across neural subdomains, resulting in limited performance. To overcome these limitations, we propose Uncertainty-guided Self-paced Cycling (UnSPC), a robust framework that synergizes DA and DG for target domain refining under an Uncertainty-guided Self-paced Pseudo-labeling (UnSPL) mechanism. To handle subdomain neural drift across domains, UNSPL is proposed to iteratively mine reliable pseudo-labeled samples with a noise-robust ranking strategy for further fine-tuning. Leveraging these high-quality samples, we introduce a novel Cycling Adaptation and Generalization (CycAG) strategy, which integrates DA and DG within an iterative cycle to progressively mitigate both global and subdomain drift. This cyclic process enables effective alignment to evolving target distributions while preserving robust and transferable representations, thereby mitigating performance degradation under long-term neural drifts. Extensive experiments on multiple neural decoding datasets demonstrate the effectiveness and robustness of UnSPC. To our knowledge, our proposed UnSPC is the first to cyclically integrate DA and DG with pseudo-labeling, paving the way toward stable long-term BMI controls.
\end{abstract}

\section{Introduction}
As a promising paradigm for direct brain-device interaction, Brain-Machine Interfaces (BMIs) have attracted growing attention due to their potential in neurorehabilitation, assistive technologies, human performance augmentation, and human-robot interaction~\cite{ICRA_Rehi,ICRA_EEG_motor,gao2025cross,rong2025improving,ICRA_VR}. Among various BMI modalities, invasive motor BMIs (iBMIs) directly record neural activity from motor cortical areas, providing high-fidelity signals that enable accurate decoding of motor intentions for effective control of prosthetic devices and robotic systems~\cite{bci1,bci2,bci3}. Over the past decade, substantial progresses in neural population recording technologies~\cite{neuronreocding1} and neural decoding algorithms~\cite{kalman,KRSRL,DyEnsemble,EvoDyEnsemble} have significantly improved decoding accuracy and expanded the practical applicability of iBMIs. However, achieving reliable long-term operation of iBMIs remains a fundamental challenge. In real-world deployments, iBMI systems often suffer from neural drift, which may be caused by electrode shifts, neuronal turnover, and neural plasticity~\cite{buttfield2006towards,Brainplasticity}. As a result, the statistical relationship between neural activity and intended movements gradually changes over time. Consequently, decoders trained on labeled $\mathtt{day\_0}$ data often generalize poorly to later sessions $\mathtt{day\_k}$, leading to progressive performance degradation and necessitating periodic recalibration.

\begin{figure}[t]
\centering
\includegraphics[width=0.99\linewidth]{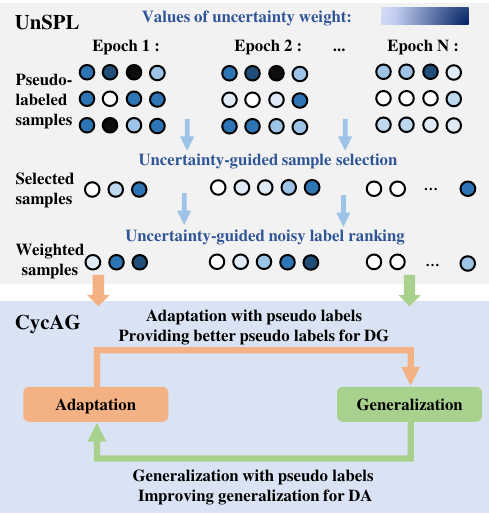}
\caption{UnSPC consists of two key modules: (i) UnSPL mines reliable pseudo-labeled samples; (ii) CycAG cyclically alternates DA and DG.}
\label{fig:UnSPC}
\end{figure}

Numerous approaches have been proposed to address neural non-stationarity, typically categorized into Domain Adaptation (DA) and Domain Generalization (DG) based methods. DA based methods maintain performance by aligning the decoder with $\mathtt{day\_k}$ statistics~\cite{adan,UAN,WDGRL,SeSA}, yet they risk negative transfer if driven by noisy features or severe distribution shifts. Conversely, DG based methods~\cite{LFDA,CAPTIVATE,CDNG,schneider2023learnable} emphasize learning invariant representations across sessions but are often overly conservative, failing to capture session-specific neural drifts by ignoring target-domain statistics. Crucially, existing methods typically treats DA and DG as isolated paradigms. In this paper, we argue that DA and DG are complementary where robust long-term neural decoding requires both adaptability to evolving neural distributions and generalization across unseen variations. Specifically, DG provides a stable initialization that shields DA from noise-induced overfitting, while DA leverages unlabeled target data to provide supervision signals that guide DG~\cite{CoDAG,zhou2024mixstyle,ghifary2016scatter}.

Furthermore, prior methods largely focus on global marginal distributions, often neglecting subdomain shifts, which are critical for fine-grained neural decoding~\cite{dsan,zhong2025deep,wei2021subdomain}. Pseudo-labeling~\cite{lee2013pseudo,cascante2021curriculum,arazo2020pseudo} offers a promising way to address this by assigning tentative labels to unlabeled target data for self-training. However, its application to iBMIs faces two unique hurdles: (i) the difficulty of confidence estimation in continuous regression (unlike discrete classification), and (ii) the risk of error propagation due to the inherent noise in neural signals. Consequently, a robust, regression-based pseudo-labeling framework for iBMIs remains an open challenge.

To this end, we propose a novel Uncertainty-guided Self-paced Cycling (UnSPC) framework, which unifies DA and DG in a single iterative learning paradigm with self-paced pseudo-labeling strategy through an uncertainty-guided mechanism. First, a Robust Pre-train Decoder (RPD) is learned on the source session through hybrid neural decoding and consistency constraints, which improve robustness to neural noise and electrode instability. To address the dual challenges of subdomain drift and label scarcity on unlabeled target domain, we introduce Uncertainty-guided Self-paced Pseudo-labeling (UnSPL) which orchestrates a robust curriculum that progressively mines high-confidence pseudo-labels, while effectively attenuating noise propagation through a rank-guided soft weighting mechanism.

Intuitively, these high-fidelity samples could serve as reliable supervision for the target domain refining, enabling a cyclic refinement process of DA and DG. Building on this, we introduce Cyclic Adaptation-Generalization (CycAG), a novel iterative optimization scheme which creates a synergistic cycle between DA and DG, progressively improving decoding performance on reliable data while extending robustness to the remaining uncertain samples. At each iteration, reliable pseudo-labeled samples are mined and weighted based on UnSPL, and further used to alternately perform DA and DG by CycAG. Note that, this alternating optimization between DA and DG enables a complementary interplay: adaptation improves target-specific performance, and generalization enhances robustness to distribution shifts, leading to a more stable and reliable decoding model under neural drift. Extensive experiments on multiple iBMI datasets demonstrate UnSPC significantly outperforms state-of-the-art DA and DG methods, achieving superior cross-session decoding performance without requiring additional labeled data from target sessions.

Our main contributions are summarized as follows:
\begin{itemize}
\item We propose UnSPC, which to our knowledge is the first unified framework integrating DA and DG with pseudo-labeling for regression-based neural decoding, effectively bridging the gap between DA and DG.
\item We introduce UnSPL to mine reliable pseudo-labeled samples from unlabeled target sessions, preventing error propagation during target domain fine-tuning.
\item We design CycAG, a cyclic optimization scheme that alternates between DA and DG, enabling complementary learning of target-specific and invariant representations.
\item Extensive validations across multiple datasets demonstrate that UnSPC significantly outperforms state-of-the-art baselines, establishing a stable computational methodology for long-term iBMI decoding.
\end{itemize}

\section{Related Work}
\textbf{Domain Adaptation (DA) based Methods.}
Supervised recalibration can effectively mitigate neural drift but is often impractical in longitudinal BMI settings due to the limited availability of labeled data~\cite{daily}. As a result, Unsupervised Domain Adaptation (UDA) has become a widely adopted strategy. Representative UDA based approaches include adversarial learning methods, such as ADAN~\cite{adan} and UAN~\cite{UAN}, which minimize the distribution discrepancy between the source ($\mathtt{day\_0}$) and target ($\mathtt{day\_k}$) domain, as well as divergence-based alignment methods, such as DA-DCF~\cite{DA-DCF} and WDGRL~\cite{WDGRL}, which align latent feature distributions across sessions. In addition, SeSA~\cite{SeSA} introduces semi-supervised DA, which leverages structural constraints to conditionally align kinematic features, thereby preserving the intrinsic manifold structure during domain adaptation.

\textbf{Domain Generalization (DG) based Methods.}
Unlike DA, DG aims to learn representations that remain robust to distribution shifts using only source-domain data. In neural decoding, DG based methods typically seek invariant neural representations that generalize across recording sessions. Some studies focus on representation learning to obtain stable neural embeddings. For example, KRSRL~\cite{KRSRL} and CDNG~\cite{CDNG} employ similarity-based objectives to enforce consistent latent representations for neural activities associated with similar movements. Some studies improve robustness through temporal modeling or ensemble strategies. For example, DyEnsemble~\cite{EvoDyEnsemble} aggregates multiple dynamic decoders to capture diverse temporal dependencies in neural signals. Some studies attempt to model invariant latent dynamics of neural populations. For example, CAPTIVATE~\cite{CAPTIVATE} and SABLE~\cite{SABLE} learn structured latent representations to stabilize neural dynamics across sessions, while LFDA~\cite{LFDA} combines neural data augmentation with contrastive learning to learn latent factors robust to recording condition changes.

\begin{figure*}[ht]
\centerline{\includegraphics[width=0.99\linewidth]{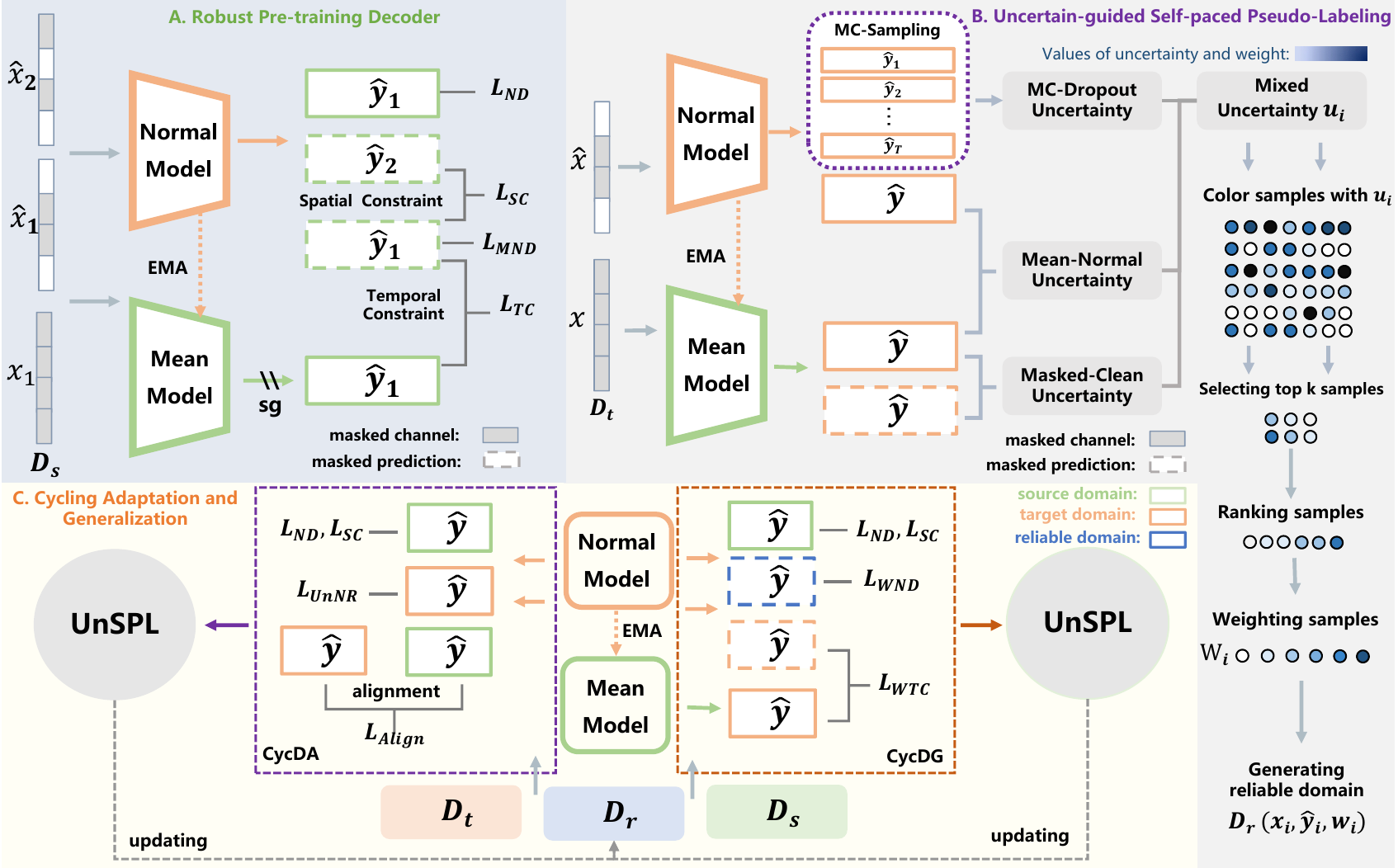}}
\caption{The overall framework of our UnSPC. A: Robust Pre-training Decoder (RPD) pre-trained on the source domain for initialization. B: Uncertainty-guided Self-paced Pseudo-labeling (UnSPL) selecting and weighing reliable samples on the target domain to form a reliable domain. C: Cyclic Adaptation-Generalization (CycAG), cyclically alternating between adaptation and generalization using reliable pseudo-labeled samples.}
\label{fig:framework}
\end{figure*}

Although the above DA and DG based methods achieve competitive performance, they overlook the benefits that integrating DA and DG could bring, limiting their performance.

\section{Method}
In this work, we propose Uncertainty-guided Self-paced Cycling (UnSPC), a framework designed to mitigate performance degradation under long-term neural drift in BCIs, as depicted in Fig.~\ref{fig:UnSPC} and Fig.~\ref{fig:framework}. The key idea of UnSPC is to progressively adapt the decoding model to evolving neural distributions while preserving robust and generalizable representations. We first introduce Robust Pre-training Decoder (RPD) as a stable backbone, which is pre-trained on the source domain via hybrid neural decoding and consistency constraints. To handle subdomain drift, we propose Uncertainty-guided Self-paced Pseudo-labeling (UnSPL), which quantifies predictive uncertainty to progressively mine high-fidelity pseudo-labeled samples from the unlabeled target domain. With these reliable samples, we introduce Cycling Adaptation-Generalization (CycAG), an iterative scheme that alternates between adaptation and generalization for target session refinement. In the following, we first formulate the neural decoding problem under session-wise distribution shifts and then introduce the proposed UnSPC and its key components.

\subsection{Problem Formulation}
We consider a neural decoding task where neural signals are used to predict corresponding behavioral variables, such as movement trajectories or intended actions. In iBMIs, neural recordings collected across different sessions often exhibit substantial distribution shifts due to neural drift and recording instability. Let $\mathcal{S}=\{(x_i^s,y_i^s)\}_{i=1}^{N_s}$ and $\mathcal{T}=\{(x_i^t,y_i^t)\}_{i=1}^{N_t}$ denote the source-session ($\mathtt{day\_0}$) dataset and a subsequent target-session ($\mathtt{day\_k}$) dataset, respectively. Here, $x_i$ and $y_i$ denote neural signals and corresponding behavioral labels respectively. The objective is to learn a neural decoding model composed of a feature extractor $f_\theta(\cdot)$ and a regression head $h_\theta(\cdot)$, i.e., $h_\theta(f_\theta(\cdot))$, that maintains high decoding accuracy on $\mathcal{T}$ despite the distribution shifts, where $\theta$ denotes the model parameters. Note that since target-session labels are unavailable during training, we assume access only to labeled $\mathtt{day\_0}$ data and unlabeled $\mathtt{day\_k}$ data.

\subsection{UnSPC Framework Overview}
On the whole, our proposed UnSPC involves two stages as follows. \textbf{(i) Source-session Pre-training:} we first train a Robust Pre-training Decoding (RPD) backbone on the labeled source session $\mathcal{S}$ as a robust initialization. \textbf{(ii) Target-session Refining:} given an unlabeled target session $\mathcal{T}$, we further refine the model through an iterative procedure; at each iteration $w$, the model first constructs a reliable pseudo-labeled set $\mathcal{R}_w$ with soft weights via UnSPL, and then $\mathcal{R}_w$ is utilized by CycDG to update the model by alternating DA and DG across successive iterations.

To stabilize training against inherent neural noise, we employ a teacher-student paradigm~\cite{meanteacher} throughout both the pre-training stage and the refining stage. The teacher parameters are updated without gradients as an exponential moving average (EMA) of the student parameters:
\begin{equation}
\hat{\theta}_w= \begin{cases}\theta_w, & \text { if } w \leq w_0 \\ \frac{w-w_0}{w-w_0+1} \cdot \hat{\theta}_{w-1}+\frac{1}{w-w_0+1} \cdot \theta_w & \text { otherwise }\end{cases},
\end{equation}
where $\hat{\theta}_w$ and $\theta_w$ denote the parameters of the teacher and student models at iteration $w$, respectively, and $w_0$ denotes the warm-up step before the EMA update starts. Since the teacher model maintains a temporally smoothed version of the student parameters, during inference the teacher model is used as the finally obtained decoder for prediction.

\subsection{Robust Pre-training Decoder}
For initialization, we propose a Robust Pre-training Decoder (RPD) backbone using labeled data during the source-session pre-training. To improve robustness against electrode instability and neural noise, we adopt a hybrid neural decoding strategy together with consistency constraints.

\subsubsection{\textbf{Hybrid Neural Decoding}}
Neural recordings are often affected by electrode instability or channel dropouts across recording sessions. To improve robustness to such perturbations, we train the decoder using hybrid neural inputs consisting of both clean and masked observations. The loss for clean neural decoding is defined as:
\begin{equation}
    \mathcal{L}_\text{CND}=\frac{1}{|\mathcal{S}|}\sum_{(x_i^s,y_i^s)\in\mathcal{S}}\|h_\theta(f_\theta(x_i^s))-y_i^s\|_2^2.
\end{equation}
To simulate electrode instability, we randomly mask $c$ neural channels in $x_i^s$ to generate a corrupted observation $\hat{x}_i^s$. The masked neural decoding loss is defined as:
\begin{equation}
    \mathcal{L}_\text{MND}=\frac{1}{|\mathcal{S}|}\sum_{(x_i^s,y_i^s)\in\mathcal{S}}\|h_\theta(f_\theta(\hat{x}_i^s))-y_i^s\|_2^2.
\end{equation}

\subsubsection{\textbf{Consistency Constraints}}
In addition to hybrid neural decoding losses, we introduce consistency constraints to stabilize representation learning and improve cross-session generalization. Neural representations corresponding to similar behavioral states should exhibit similar geometric relationships. To preserve this structure, we enforce alignment between pairwise distances in the neural feature space and those in the behavioral space. The feature-label consistency loss is defined as:
\begin{equation}
    \mathcal{L}_\text{FLC}=\frac{1}{|\mathcal{S}|^2}\sum_{(x_i^s,y_i^s)\in\mathcal{S}}\sum_{(x_j^s,y_j^s)\in\mathcal{S}} \delta_{ij},
\end{equation}
and
\begin{equation}
\delta_{ij}=\Big(\text{BN}(\|f_\theta(x_i^s)-f_\theta(x_j^s)\|_2^2)-\text{BN}(\|y_i^s-y_j^s\|_2^2)\Big)^2,
\end{equation}
where $\text{BN}(\cdot)$ denotes the batch normalization operator applied to stabilize the scale of pairwise distances. Besides, to improve robustness to corrupted inputs, we enforce consistency between predictions of the teacher model on clean signals and predictions of the student model on masked signals. The teacher-student consistency loss is defined as:
\begin{equation}
    \mathcal{L}_\text{TSC}=\frac{1}{|\mathcal{S}|}\sum_{(x_i^s,y_i^s)\in\mathcal{S}}
    \|h_{\hat{\theta}}(f_{\hat{\theta}}(x_i^s))-h_\theta(f_\theta(\hat{x}_i^s))\|_2^2.
\end{equation}

\subsubsection{\textbf{Overall Loss}}
The final objective for source-session pre-training combines the hybrid decoding losses and the consistency constraints:
\begin{equation}
    \mathcal{L}_\text{PRE}=\alpha\cdot(\mathcal{L}_\text{CND}+\mathcal{L}_\text{FLC})+\mathcal{L}_\text{MND}+\mathcal{L}_\text{TSC},
\end{equation}
where $\alpha$ controls the trade-off between the losses involving both clean and masked neural signals and the losses involving only clean neural signals. By jointly optimizing these objectives, the decoder learns noise-robust neural representations that preserve the geometric relationship between neural activity and behavioral states, providing a reliable initialization for subsequent target-session refining.

\subsection{Uncertainty-guided Self-paced Pseudo-labeling}
Despite the above initialization, significant distribution shifts between $\mathcal{S}$ and $\mathcal{T}$ necessitate further fine-tuning. 
While conventional UDA aligns global feature distributions~\cite{DA-DCF,SeSA}, it often fails to enforce the fine-grained manifold structure required for high-precision regression. Pseudo-labeling~\cite{lee2013pseudo} offers instance-level supervision but faces two impediments in iBMIs: the lack of a natural confidence metric for regression, and the risk of error propagation from noisy neural signals. To surmount these, we propose Uncertainty-guided Self-paced Pseudo-labeling (Fig.~\ref{fig:framework}-B) to select and weigh pseudo-labeled samples on $\mathcal{T}$.

Given the unlabeled target-session data, we first perform Reliable Sample Selection (RSS), estimating the reliability of model predictions and constructing a set of pseudo-labeled samples. Specifically, for each target sample $x_i^t$, we obtain its pseudo-label $\hat{y}_i^t$ using the teacher model:
\begin{equation}
    \hat{y}_i^t = h_{\hat{\theta}}(f_{\hat{\theta}}(x_i^t)).
\end{equation}
To assess the reliability of pseudo-labels, we estimate the prediction uncertainty from two perspectives. First, we compute the Monte-Carlo (MC) Dropout uncertainty by performing $M$ stochastic forward passes with dropout activated at inference time:
\begin{equation}
    u_i^{\text{MC}} = \frac{1}{M}\sum_{j=1}^{M}\|\hat{y}_{i,j}^t - \mathbb{E}[\hat{y}_i^t]\|_2^2.
\end{equation}
Second, we measure the inconsistency (IC) between different model predictions:
\begin{equation}
    u_i^{\text{IC}} = \|\hat{y}_i^t-h_{\hat{\theta}}(f_{\hat{\theta}}(\hat{x}_i^t))\|_2^2 + \|\hat{y}_i^t-h_{\theta}(f_{\theta}(x_i^t))\|_2^2,
\end{equation}
where $\hat{x}_i^t$ denotes the masked version of $x_i^t$. The final uncertainty is computed as:
\begin{equation}
    u_i = \text{BN}(u_i^{\text{MC}}) + \text{BN}(u_i^{\text{IC}}).
\end{equation}
Based on the uncertainty values, we adopt a self-paced strategy to progressively select reliable samples. We first define a candidate reliable set parameterized by a threshold:
\begin{equation}
    \mathcal{R}(\tau) = \{(x_i^t, \hat{y}_i^t) \mid u_i \leq \tau \}.
\end{equation}
At iteration $w$, the threshold $\tau_w$ is then chosen as the smallest value that ensures a desired proportion of selected samples:
\begin{equation}
    \tau_w = \min\{ \tau \;|\; \frac{|\mathcal{R}(\tau)|}{|\mathcal{T}|} \ge \gamma_0 + w \cdot \Delta\gamma\},
\end{equation}
where $\gamma_0$ controls the initial sampling ratio and $\Delta\gamma$ defines the incremental increase at each iteration. 

Accordingly, the reliable set at iteration $w$ is given by $\mathcal{R}_w = \mathcal{R}(\tau_w)$. This formulation ensures that the model first focuses on highly reliable samples and gradually incorporates more uncertain ones as training progresses. Samples in $\mathcal{R}_w$ are ranked in ascending order based on their uncertainty values. Let $r_i$ denote the rank of sample $x_i^t$. To further mitigate noise in pseudo-labels, we perform Reliable Sample Weighing (RSW), assigning each sample a weight based on its uncertainty ranking as follows:
\begin{equation}
    z_i = \exp(-\beta \cdot \frac{r_i}{|\mathcal{R}_w|}),
\end{equation}
This weighing mechanism allows more reliable samples to contribute more during training.

\subsection{Cyclic Adaptation-Generalization for Refinement}
By treating these reliable samples as newly labeled data, we move beyond considering adaptation and generalization as independent processes, and instead leverage them to jointly enhance both. This motivates the proposed Cyclic Adaptation-Generalization (CycAG), which alternates between DA and DG in a cyclic manner (Fig.~\ref{fig:framework}-C).

\subsubsection{\textbf{Domain Adaptation}}
In this step, we leverage the reliable pseudo-labeled set $\mathcal{R}_w$ to align the feature distributions between the source and target domains while improving prediction accuracy. First, we adopt a feature alignment loss based on Maximum Mean Discrepancy (MMD):
\begin{equation}
    \mathcal{L}_\text{MMD} = \left\| \frac{1}{|\mathcal{S}|}\sum_{(x_i^s,y_i^s) \in \mathcal{S}} f_\theta(x_i^s) - \frac{1}{|\mathcal{T}|}\sum_{(x_i^t,y_i^t) \in \mathcal{T}} f_\theta(x_i^t) \right\|_{\mathcal{H}}^2,
\end{equation}
where $\mathcal{H}$ denotes a reproducing kernel Hilbert space. Then, we utilize the pseudo-labels and the assigned weights for weighted supervised learning (WSL) on the reliable set:
\begin{equation}
    \mathcal{L}_\text{WSL} = \frac{1}{|\mathcal{R}_w|} \sum_{(x_i^t,\hat{y}_i^t)\in\mathcal{R}_w} z_i \cdot \|h_\theta(f_\theta(x_i^t)) - \hat{y}_i^t\|_2^2.
\end{equation}
The overall loss for the DA step is:
\begin{equation}
    \mathcal{L}_\text{DA} = \alpha\cdot(\mathcal{L}_\text{CND} + \mathcal{L}_\text{FLC}) + \beta\cdot\mathcal{L}_\text{WSL} + \eta\cdot\mathcal{L}_\text{MMD}.
\end{equation}
This step focuses on improving target-domain performance by aligning global distributions and refining predictions using reliable pseudo-labels.

\subsubsection{\textbf{Domain Generalization}}
Complementary to DA, this step aims to improve the robustness of the model to unseen variations in neural signals. Instead of directly fitting clean inputs, we encourage the model to learn invariant representations under corrupted inputs. We first introduce a teacher-student consistency loss over the target domain (TSCT):
\begin{equation}
    \mathcal{L}_\text{TSCT} = \frac{1}{|\mathcal{T}|} \sum_{(x_i^t,y_i^t) \in \mathcal{T}} \|h_{\hat{\theta}}(f_{\hat{\theta}}(x_i^t))-h_\theta(f_\theta(\hat{x}_i^t))\|_2^2.
\end{equation}
Then, we define a weighted masked neural decoding (WMND) loss on the reliable samples:
\begin{equation}
    \mathcal{L}_\text{WMND} = \frac{1}{|\mathcal{R}_w|} \sum_{(x_i^t,\hat{y}_i^t)\in\mathcal{R}_w} z_i \cdot \|h_\theta(f_\theta(\hat{x}_i^t)) - \hat{y}_i^t\|_2^2.
\end{equation}
The overall loss for the DG step is:
\begin{equation}
    \mathcal{L}_\text{DG} = \alpha\cdot(\mathcal{L}_\text{CND} + \mathcal{L}_\text{FLC}) + \beta\cdot \mathcal{L}_\text{WMND} + \phi\cdot\mathcal{L}_\text{TSCT}.
\end{equation}
By enforcing consistency under input perturbations and leveraging weighted pseudo-label supervision, this step enhances the model's generalizability to unseen distributions.

\begin{figure*}[t]
\centerline{\includegraphics[width=0.99\linewidth]{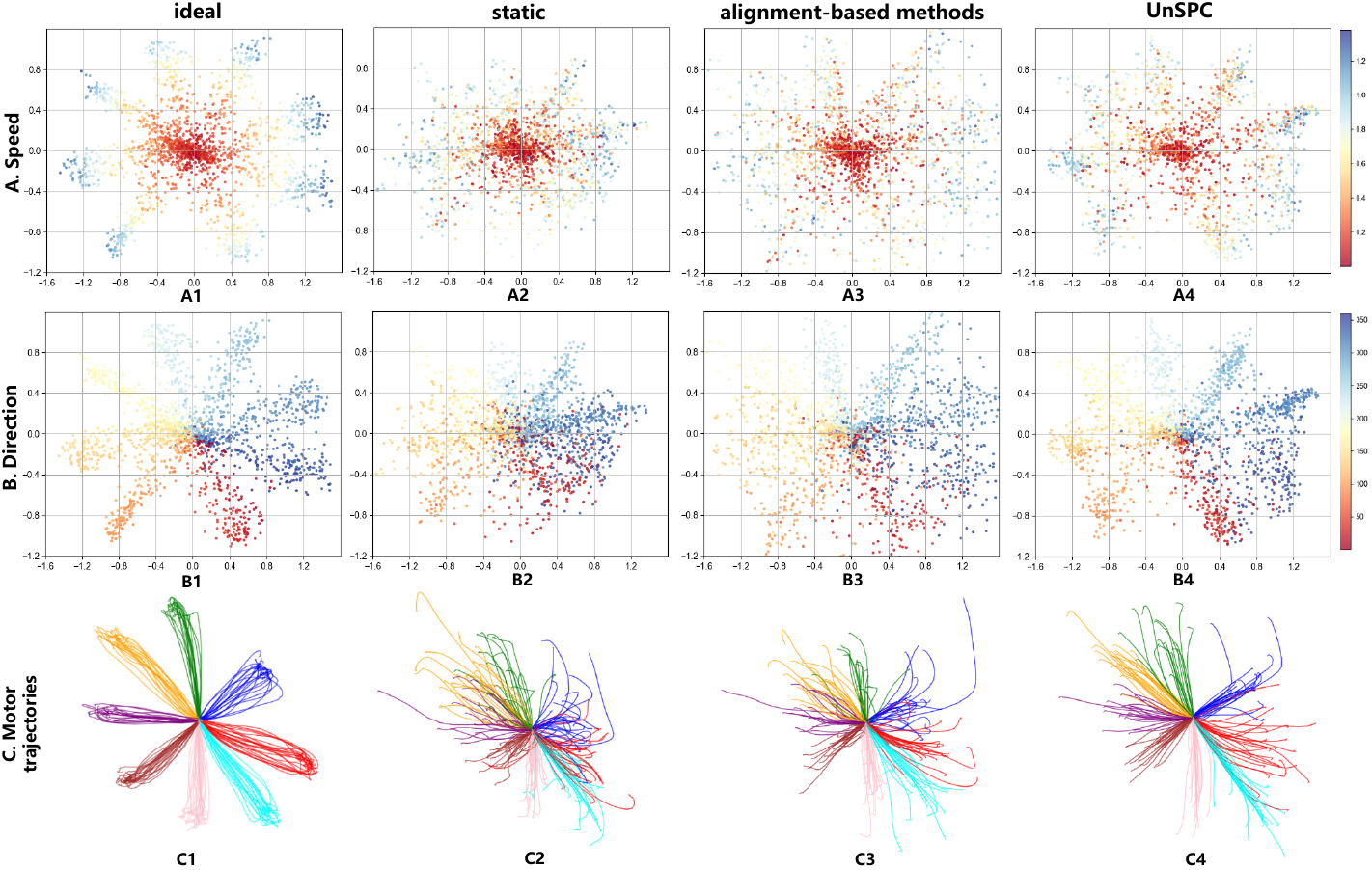}}
\caption{Feature distributions and motor trajectories of ideal (supervised), static (uncalibrated), alignment-based method (WDGRL) and our UnSPC on C0--C1. A and B: Feature distributions under different alignments colored by values of speed and direction. C: Motor trajectories colored by trial direction.}
\label{fig:exp-main}
\end{figure*}

\begin{table*}[t]
    \centering
    \caption{Recalibration performance comparison (bold for the best and underline for the second-best)}
    \begin{tabularx}{0.99\linewidth}{cX<{\centering}X<{\centering}X<{\centering}X<{\centering}X<{\centering}X<{\centering}X<{\centering}X<{\centering}X<{\centering}X<{\centering}X<{\centering}X<{\centering}X<{\centering}X<{\centering}}
    \toprule
    \multirow{4}{*}{Dataset} & \multicolumn{14}{c}{Method} \\
    \cmidrule(lr){2-15}& \multicolumn{2}{c}{Static} & \multicolumn{2}{c}{LFDA}  & \multicolumn{2}{c}{DA-DCF*} & \multicolumn{2}{c}{UAN} & \multicolumn{2}{c}{WDGRL*} & \multicolumn{2}{c}{SeSA} & \multicolumn{2}{c}{\textbf{UnSPC}}\\
    \cmidrule(lr){2-3}\cmidrule(lr){4-5}\cmidrule(lr){6-7}\cmidrule(lr){8-9}\cmidrule(lr){10-11}\cmidrule(lr){12-13}\cmidrule(lr){14-15} & $CC$ & $R^2$ & $CC$ & $R^2$ & $CC$ & $R^2$ & $CC$ & $R^2$ & $CC$ & $R^2$ & $CC$ & $R^2$ & $CC$ & $R^2$\\
	\midrule		
	 C1 &  $84.78$ & $59.24$ &  $87.91$ & $72.35$  &   $87.35$ & $62.03$ &  $84.35$ & $64.79$ &  $\underline{86.92}$ & $\underline{72.45}$ &  $84.39$ & $69.43 $   &  $\textbf{88.70}$ & $ \textbf{78.40} $\\  
	 C2 &  $65.94$ & $40.55$ & $74.58$ & $55.12$  &   $75.12$ & $51.64$ &  $77.91$ & $55.57$ &  $78.82$ & $65.92$  &  $\underline{84.59}$ & $\underline{66.30}$    &  $\textbf{88.91}$ & $\textbf{78.77}$ \\  
	 C3 &  $57.43$ & $30.23$ & $68.14$ & $41.21$  &    $64.89$ & $40.11$ &  $70.98$ & $46.28$ & $80.09$ & $60.03$  &  $\underline{82.15}$ & $\underline{65.35}$  & $\textbf{84.94}$ & $\textbf{71.64} $ \\ 
    \midrule 		
	 J1 &  $81.89$ & $66.41$ & $\underline{84.39}$ & $\underline{70.66}$  &  $83.09 $ & $67.62$ &  $81.19$ & $65.04$ &  $83.26$ & $68.05$ & $83.14$ & $70.66$  &  $\textbf{85.42}$ &  $\textbf{72.08}$ \\
	 J2 &  $74.71$ & $56.24$ & $82.28$ & $65.30$  &   $77.75$ & $59.01$ &  $79.74$ & $63.61$ & $78.32$ & $64.98$ &  $\underline{82.45}$ & $\underline{65.75}$  & $\textbf{83.21}$ &  $\textbf{68.13}$ \\
	 J3 &  $77.48$ & $60.95$ & $80.86$ & $58.43$  &  $75.84$ & $58.55$ &  $77.82$ & $59.63$ &  $78.15$ & $62.33$ & $\underline{81.51}$ & $\underline{67.73}$  & $\textbf{83.38}$ &  $\textbf{68.33} $ \\
    \midrule 		
	 M1 &  $77.26$ & $64.82$ & $82.22$ & $64.77$  &  $77.89$ & $65.61$ &  $78.21$ & $65.72$ & $78.52$ & $65.49$ &   $\underline{78.09}$ & $\underline{66.17}$ &   $\textbf{82.25}$ &  $\textbf{67.49} $ \\
	 M2 &  $67.66$ & $49.17$ & $74.10$ & $54.89$  &  $67.84$ & $50.15$ &  $67.94$ & $51.12$ &  $71.28$ & $55.58$ &  $\underline{75.10}$ & $\underline{60.78}$ &   $\textbf{79.66}$ &  $\textbf{62.99} $ \\
	 M3 &  $58.45$ & $30.12$ & $65.24$ & $42.30$  & $42.31$ & $30.43$ &  $49.51$ & $36.79$ & $55.21$ & $40.56$ & $\textbf{71.57}$ & $\textbf{54.24}$  &   $\underline{68.61}$ &  $\underline{46.99}$ \\
    \bottomrule
    \end{tabularx}
    \label{tab:2} 
\end{table*}

\section{Experiments}
In this section, we first describe the experimental settings, and then present comparisons with state-of-the-art methods across datasets, followed by ablation and robustness studies to comprehensively evaluate UnSPC.

\subsection{Experimental Settings}
\subsubsection{\textbf{Datasets}}
We validate UnSPC on three public NHP datasets (Chewie, Mihili, and Jango)~\cite{UAN}. Subjects performed upper-limb tasks including center-out (CO), random-target (RT), and isometric wrist movements (ISO), with neural activity recorded from the primary motor cortex (M1) using 96-channel Utah arrays. Multiunit threshold crossings were extracted, randomly split, and aligned to the central $10\%$ to $90\%$ of each trial (from "go cue" to "trial end"). Firing rates were computed using $50$ ms binning, followed by Gaussian smoothing ($\sigma = 100$ ms) and z-score normalization. For long-term stability evaluation, the first session was used as the labeled source session, and subsequent sessions as unlabeled target sessions.

\subsubsection{\textbf{Implementation Details}}
The feature extractor is a lightweight 3-layer MLP with ReLU activations and $p=0.25$ dropout, followed by a linear regression head. We optimize the model using Adam with $learning\_rate=5\times10^{-3}$, $beta=(0.9, 0.999)$, and $weight\_decay=5\times10^{-4}$. The batch size is $128$. Hyperparameters are fixed at $\beta=0.1$, $\eta=0.2$, and $\phi=0.5$ across all datasets. To account for subject variability, the source preservation weight is set to $\alpha=1.0$ for Chewie and Mihili, and $\alpha=0.1$ for Jango. Results are reported as the mean of five independent runs with different random seeds.

\subsubsection{\textbf{Metrics}}
Decoding performance is assessed using three metrics: Pearson Correlation Coefficient ($CC$), Coefficient of Determination ($R^2$), and Mean Squared Error ($MSE$).

\subsection{Recalibration Performance Comparison}
We evaluate the recalibration performance through qualitative and quantitative experiments across multiple datasets. Data from four days per subject were used: C0--C3 (Chewie), J0--J3 (Jango), and M0--M3 (Mihili). C0/J0/M0 serve as source domains, and the remaining sessions correspond to short ($\sim$1 day), medium ($\sim$1 month), and long ($\sim$3 months) time spans. ``*'' denotes our reproduced results.

\subsubsection{\textbf{Qualitative Results}}
From the qualitative comparisons of feature distributions (Fig.~\ref{fig:exp-main}-A and B) we can reveal that:
{\textbf{(i)}} An ideal feature distribution exhibits a radial shape and maintains label-feature consistency.
{\textbf{(ii)}} The feature distributions of the static decoder show significant deviations.
{\textbf{(iii)}} Previous methods perform feature alignment, which reduces the deviations to some extent.
{\textbf{(iv)}} Our UnSPC achieves nearly ideal feature distributions.
A similar conclusion can be drawn from the motor trajectories in Fig.~\ref{fig:exp-main}-C, where the trajectories decoded by our method are closer to the ideal motion trajectories, demonstrating the effectiveness and superiority of UnSPC from a qualitative perspective.

\subsubsection{\textbf{Quantitative results}}
As shown in Table~\ref{tab:2}, quantitative results show that the static Decoder degrades significantly without recalibration. DA based methods (DA-DCF, UAN, WDGRL) partially mitigate this via feature alignment, and the DG based method (LFDA) performs comparably, yet all remain limited by the absence of labels. The semi-supervised method SeSA improves over most unsupervised methods but requires a few labeled target data. In contrast, UnSPC substantially outperforms the static decoder baseline and achieves state-of-the-art performance among DA and DG based methods across all datasets, surpassing SeSA on most datasets despite being fully unsupervised.

\subsection{Ablation Study}
To validate the effectiveness of each component in UnSPC, we conduct ablation experiments with C0 as the source domain and C1 as the target domain (Table~\ref{tab:3}).

\begin{table}[t]
\centering
    \caption{Ablation results of UnSPC}
    \begin{tabularx}{0.99\linewidth}{lX<{\centering}X<{\centering}X<{\centering}}
    \toprule
    Configuration & $CC$ & $R^2$ & $MSE$ \\
    \midrule
 	Baseline   & $84.78$ &  $59.24$ & $.03668$  \\
  	Baseline+RPD  & $87.22$ &  $73.04$ &$.02504$  \\	
    \midrule 
	UnSPC w/o RSS  & $88.56$ &  $77.46$ &  $.02256$ \\
    UnSPC w/o RSW  & $88.37$ & $\underline{77.80}$ & $\underline{.02225}$ \\
    UnSPC w/o UnSPL & $86.78$ & $73.30$ & $.02688$ \\
    \midrule
    UnSPC w/o CycDA & $\underline{88.57}$ & $76.36$ & $.02379$   \\
 	UnSPC w/o CycDG & $87.94$ & $77.01$ & $.02305$   \\
    UnSPC w/o Cyc & $88.52$ & $ 77.69$ & $.02263$ \\
    \midrule
    UnSPC  & $\textbf{88.70}$ &  $\textbf{78.40}$ & $\textbf{.02168}$ \\
    \bottomrule
	\end{tabularx}
    \label{tab:3}
\end{table}

\subsubsection{\textbf{RPD is a strong backbone}}
Directly applying a source-trained neural decoder to the target domain results in a significant performance drop, highlighting neural signal shifts across days. Incorporating RPD substantially mitigates this drop, confirming its role as a strong backbone.

\subsubsection{\textbf{UnSPL is effective and critical}}
Removing either RSS or RSW leads to performance degradation, demonstrating their significance. Without UnSPL (i.e., applying pseudo-labels directly) performance drops severely (lower than Baseline+RPD), confirming that pseudo-label noise can be detrimental and UnSPL is essential for effective recalibration.

\subsubsection{\textbf{CycAG can improve performance}}
When UnSPC is trained without CycDA and CycDG, relying only on RPD and UnSPL, performance drops compared to using CycDA or CycDG individually, confirming that CycDA improves adaptation on unlabeled data while CycDG enhances generalization. Combining DA and DG without cycling yields competitive results, but the full CycAG framework achieves the best performance, demonstrating the effectiveness of iterative cycling. These findings highlight that the synergistic integration of UnSPL and CycAG leads to optimal recalibration, establishing UnSPC as a state-of-the-art method for neural decoding.

\subsection{Robustness Analysis}
\subsubsection{\textbf{Sensitivity Analysis of Hyperparameters}}
We assess the sensitivity of $\alpha$, $\beta$, $\eta$, and $\phi$, keeping other parameters at default values. As shown in Table~\ref{tab:sensitivity}, $\alpha$ should be sufficiently large to preserve baseline performance, $\beta$ moderate to leverage pseudo-labels without amplifying noise, while $\eta$ and $\phi$ have minor effects but can stabilize training. Overall, within a reasonably wide range, UnSPC demonstrates low sensitivity to these hyperparameters, indicating robust performance across different settings.

\subsubsection{\textbf{Robustness on Human Dataset}}
We further evaluated the robustness of UnSPC on a human robotic effector dataset from FALCON~\cite{focal}, where the source domain consists of recordings on Date 01-01 and the participant performed a virtual arm reaching task. Table~\ref{human} reports performance across three subsequent target days. Compared with the static decoder and SOTA methods, UnSPC consistently achieves the highest performance, demonstrating its effectiveness and robustness in real-world human BMI scenarios. These results indicate that our method can reliably handle cross-day neural variability and maintain stable decoding without access to labeled target data.

\begin{table}[t]
\centering
\caption{Sensitivity of Hyperparameters on C1}
\begin{tabularx}{0.99\linewidth}{cX<{\centering}X<{\centering}}
\toprule
Parameter & Tested Values & $R^2$ \\
\midrule
$\alpha$ & 0.1, 0.5, 1.0, 2.0 & 73.0, 76.5, 78.4, 78.2 \\
$\beta$ & 0.05, 0.1, 0.2, 0.5 & 77.5, 78.4, 77.6, 75.8 \\
$\eta$ & 0.05, 0.2, 0.5 & 78.0, 78.4, 78.3 \\
$\phi$ & 0.1, 0.5, 1.0 & 77.9, 78.4, 78.5 \\
\hline
\end{tabularx}
\label{tab:sensitivity}
\end{table}
\begin{table}[t]
\centering
    \caption{Robustness on Human Dataset}
	\begin{tabularx}{0.99\linewidth}{cX<{\centering}X<{\centering}X<{\centering}X<{\centering}}
    \toprule
	 Target Domain & Static & UAN & WDGRL & UnSPC \\
    \midrule 
 	01-08 & $10.43$ & $12.01$ & $\underline{12.71}$ & $\textbf{14.58}$ \\
    01-20 & $9.14$ & $11.16$ & $\underline{12.39}$ & $\textbf{14.17}$ \\	
  	02-09 & $8.76$ & $10.98$ & $\underline{11.74}$ & $\textbf{13.91}$ \\	
	\bottomrule 
	\end{tabularx}
    \label{human}
\end{table}

\section{Conclusion}
In this paper, we propose UnSPC, a novel framework for long-term iBMIs that effectively integrates DA and DG with pseudo-labeling to address subdomain neural drift. Extensive experiments across multiple datasets demonstrate that UnSPC achieves superior performance compared to existing methods, highlighting its robustness and practical potential for human-robot interaction and control in robotic systems.

\section{Acknowledgment}
This work was supported in part by the National Natural Science Foundation of China under Grant (U25D9015, 62336007), in part by the Starry Night Science Fund of the Zhejiang University Shanghai Institute for Advanced Study under Grant SN-ZJU-SIAS-002, in part by the Fundamental Research Funds for the Central Universities. The use of Gemini are acknowledged by authors for text polishing for this manuscript.

\bibliographystyle{IEEEtran}
\bibliography{refs}

@article{bci1,
  title={Neuronal ensemble control of prosthetic devices by a human with tetraplegia},
  author={Hochberg, Leigh R and Serruya, Mijail D and Friehs, Gerhard M and Mukand, Jon A and Saleh, Maryam and Caplan, Abraham H and Branner, Almut and Chen, David and Penn, Richard D and Donoghue, John P},
  journal={Nature},
  volume={442},
  number={7099},
  pages={164--171},
  year={2006},
  publisher={Nature Publishing Group UK London}
}

@article{bci2,
  title={Reach and grasp by people with tetraplegia using a neurally controlled robotic arm},
  author={Hochberg, Leigh R and Bacher, Daniel and Jarosiewicz, Beata and Masse, Nicolas Y and Simeral, John D and Vogel, Joern and Haddadin, Sami and Liu, Jie and Cash, Sydney S and Van Der Smagt, Patrick and others},
  journal={Nature},
  volume={485},
  number={7398},
  pages={372--375},
  year={2012},
  publisher={Nature Publishing Group UK London}
}

@article{bci3,
  title={High performance communication by people with paralysis using an intracortical brain-computer interface},
  author={Pandarinath, Chethan and Nuyujukian, Paul and Blabe, Christine H and Sorice, Brittany L and Saab, Jad and Willett, Francis R and Hochberg, Leigh R and Shenoy, Krishna V and Henderson, Jaimie M},
  journal={elife},
  volume={6},
  pages={e18554},
  year={2017},
  publisher={eLife Sciences Publications, Ltd}
}

@article{kalman,
  title={Ensemble recordings of human subcortical neurons as a source of motor control signals for a brain-machine interface},
  author={Patil, Parag G and Carmena, Jose M and Nicolelis, Miguel AL and Turner, Dennis A},
  journal={Neurosurgery},
  volume={55},
  number={1},
  pages={27--38},
  year={2004},
  publisher={LWW}
}

@article{DyEnsemble,
  title={Dynamic ensemble bayesian filter for robust control of a human brain-machine interface},
  author={Qi, Yu and Zhu, Xinyun and Xu, Kedi and Ren, Feixiao and Jiang, Hongjie and Zhu, Junming and Zhang, Jianmin and Pan, Gang and Wang, Yueming},
  journal={IEEE Transactions on Biomedical Engineering},
  volume={69},
  number={12},
  pages={3825--3835},
  year={2022},
  publisher={IEEE}
}

@article{EvoDyEnsemble,
  title={Tracking Functional Changes in Nonstationary Signals with Evolutionary Ensemble Bayesian Model for Robust Neural Decoding},
  author={Zhu, Xinyun and Qi, Yu and Pan, Gang and Wang, Yueming},
  journal={Advances in Neural Information Processing Systems},
  volume={35},
  pages={22576--22588},
  year={2022}
}

@article{KRSRL,
  title={Robust neural decoding by kernel regression with Siamese representation learning},
  author={Li, Yangang and Qi, Yu and Wang, Yiwen and Wang, Yueming and Xu, Kedi and Pan, Gang},
  journal={Journal of Neural Engineering},
  volume={18},
  number={5},
  pages={056062},
  year={2021},
  publisher={IOP Publishing}
}

@article{neuronreocding1,
  title={Using multi-neuron population recordings for neural prosthetics},
  author={Chapin, John K},
  journal={Nature neuroscience},
  volume={7},
  number={5},
  pages={452--455},
  year={2004},
  publisher={Nature Publishing Group US New York}
}

@article{Brainplasticity,
  title={Brain plasticity and reorganization before, during, and after glioma resection},
  author={Duffau, Hugues},
  journal={Glioblastoma},
  pages={225--236},
  year={2016},
  publisher={Elsevier Inc}
}

@article{daily,
  title={Restoration of reaching and grasping movements through brain-controlled muscle stimulation in a person with tetraplegia: a proof-of-concept demonstration},
  author={Ajiboye, A Bolu and Willett, Francis R and Young, Daniel R and Memberg, William D and Murphy, Brian A and Miller, Jonathan P and Walter, Benjamin L and Sweet, Jennifer A and Hoyen, Harry A and Keith, Michael W and others},
  journal={The Lancet},
  volume={389},
  number={10081},
  pages={1821--1830},
  year={2017},
  publisher={Elsevier}
}

@article{meanteacher,
  title={Mean teachers are better role models: Weight-averaged consistency targets improve semi-supervised deep learning results},
  author={Tarvainen, Antti and Valpola, Harri},
  journal={Advances in neural information processing systems},
  volume={30},
  year={2017}
}

@inproceedings{adan,
  author       = {Ali Farshchian and
                  Juan Alvaro Gallego and
                  Joseph Paul Cohen and
                  Yoshua Bengio and
                  Lee E. Miller and
                  Sara A. Solla},
  title        = {Adversarial Domain Adaptation for Stable Brain-Machine Interfaces},
  booktitle    = {7th International Conference on Learning Representations, {ICLR} 2019,
                  New Orleans, LA, USA, May 6-9, 2019},
  publisher    = {OpenReview.net},
  year         = {2019},
  bibsource    = {dblp computer science bibliography, https://dblp.org}
}

@article{UAN,
  title={Using adversarial networks to extend brain computer interface decoding accuracy over time},
  author={Ma, Xuan and Rizzoglio, Fabio and Bodkin, Kevin L and Perreault, Eric and Miller, Lee E and Kennedy, Ann},
  journal={elife},
  volume={12},
  pages={e84296},
  year={2023},
  publisher={eLife Sciences Publications Limited}
}

@article{WDGRL,
  title={Learning robust features from nonstationary brain signals by multiscale domain adaptation networks for seizure prediction},
  author={Qi, Yu and Ding, Ling and Wang, Yueming and Pan, Gang},
  journal={IEEE Transactions on Cognitive and Developmental Systems},
  volume={14},
  number={3},
  pages={1208--1216},
  year={2021},
  publisher={IEEE}
}

@article{DA-DCF,
  title={Decoder calibration framework for intracortical brain-computer interface system via domain adaptation},
  author={Dong, Yuanrui and Hu, Dingyin and Wang, Shirong and He, Jiping},
  journal={Biomedical Signal Processing and Control},
  volume={81},
  pages={104453},
  year={2023},
  publisher={Elsevier}
}

@article{dsan,
  title={Deep subdomain adaptation network for image classification},
  author={Zhu, Yongchun and Zhuang, Fuzhen and Wang, Jindong and Ke, Guolin and Chen, Jingwu and Bian, Jiang and Xiong, Hui and He, Qing},
  journal={IEEE transactions on neural networks and learning systems},
  volume={32},
  number={4},
  pages={1713--1722},
  year={2020},
  publisher={IEEE}
}

@inproceedings{lee2013pseudo,
  title={Pseudo-label: The simple and efficient semi-supervised learning method for deep neural networks},
  author={Lee, Dong-Hyun and others},
  booktitle={Workshop on challenges in representation learning, ICML},
  volume={3},
  number={2},
  pages={896},
  year={2013},
  organization={Atlanta}
}

@article{LFDA,
  title={Increasing Robustness of Intracortical Brain-Computer Interfaces for Recording Condition Changes via Data Augmentation},
  author={Yang, Shih-Hung and Huang, Chun-Jui and Huang, Jhih-Siang},
  journal={Computer Methods and Programs in Biomedicine},
  volume={251},
  pages={108208},
  year={2024},
  publisher={Elsevier}
}

@inproceedings{CAPTIVATE,
  title={Capturing cross-session neural population variability through self-supervised identification of consistent neuron ensembles},
  author={Jude, Justin and Perich, Matthew G and Miller, Lee E and Hennig, Matthias H},
  booktitle={NeurIPS Workshop on Symmetry and Geometry in Neural Representations},
  pages={234--257},
  year={2023},
  organization={PMLR}
}

@inproceedings{SABLE,
  title={Robust alignment of cross-session recordings of neural population activity by behaviour via unsupervised domain adaptation},
  author={Jude, Justin and Perich, Matthew and Miller, Lee and Hennig, Matthias},
  booktitle={International Conference on Machine Learning},
  pages={10462--10475},
  year={2022},
  organization={PMLR}
}

@INPROCEEDINGS{SeSA,
  author={Wei, Jiyu and Rong, Dazhong and Zhu, Xinyun and He, Qinming and Wang, Yueming},
  booktitle={2024 IEEE International Conference on Bioinformatics and Biomedicine (BIBM)}, 
  title={Speed-enhanced Subdomain Alignment for Long-term Stable Neural Decoding in Brain-computer Interfaces}, 
  year={2024},
  volume={},
  number={},
  pages={3832-3837},
  doi={10.1109/BIBM62325.2024.10822151}}

@inproceedings{ICRA_VR,
  author       = {Alexander Thomas and
                  Jianan Chen and
                  Anna Hella{-}Szabo and
                  Merlin Angel Kelly and
                  Tom Carlson},
  title        = {High stimuli virtual reality training for a brain controlled robotic
                  wheelchair},
  booktitle    = {{IEEE} International Conference on Robotics and Automation,
                  2024, Yokohama, Japan, 2024},
  pages        = {11305--11311},
  year         = {2024},
  doi          = {10.1109/ICRA57147.2024.10610636},
  bibsource    = {dblp computer science bibliography, https://dblp.org}
}

@inproceedings{ICRA_EEG_motor,
  author       = {Ho Jin Choi and
                  Satyajeet Das and
                  Shaoting Peng and
                  Ruzena Bajcsy and
                  Nadia Figueroa},
  title        = {On the Feasibility of EEG-based Motor Intention Detection for Real-Time
                  Robot Assistive Control},
  booktitle    = {{IEEE} International Conference on Robotics and Automation,
                  2024, Yokohama, Japan, May 13-17, 2024},
  pages        = {5592--5599},
  publisher    = {{IEEE}},
  year         = {2024},
  bibsource    = {dblp computer science bibliography, https://dblp.org}
}

@inproceedings{ICRA_Rehi,
  author       = {Kecheng Shi and
                  Rui Huang and
                  Fengjun Mu and
                  Zhinan Peng and
                  Ke Huang and
                  Yizhe Qin and
                  Xiao Yang and
                  Hong Cheng},
  title        = {A Novel Multimodal Human-Exoskeleton Interface Based on {EEG} and
                  sEMG Activity for Rehabilitation Training},
  booktitle    = {2022 International Conference on Robotics and Automation, 2022,
                   PA, USA, 2022},
  pages        = {8076--8082},
  publisher    = {{IEEE}},
  year         = {2022},
  bibsource    = {dblp computer science bibliography, https://dblp.org}
}

@inproceedings{CoDAG,
  title={Complementary domain adaptation and generalization for unsupervised continual domain shift learning},
  author={Cho, Wonguk and Park, Jinha and Kim, Taesup},
  booktitle={Proceedings of the IEEE/CVF International Conference on Computer Vision},
  pages={11442--11452},
  year={2023}
}

@article{focal,
  title={Few-shot algorithms for consistent neural decoding (falcon) benchmark},
  author={Karpowicz, Brianna M and Ye, Joel and Fan, Chaofei and Tostado-Marcos, Pablo and Rizzoglio, Fabio and Washington, Clay and Scodeler, Thiago and de Lucena, Diogo and Nason-Tomaszewski, Samuel R and Mender, Matthew J and others},
  journal={Advances in Neural Information Processing Systems},
  volume={37},
  pages={76578--76615},
  year={2024}
}

@INPROCEEDINGS{CDNG,
  author={Wei, Jiyu and Rong, Dazhong and Hong, Di and Zhang, Zhanjie and Zhu, Xinyun and He, Qinming and Wang, Yueming},
  booktitle={2025 IEEE International Conference on Bioinformatics and Biomedicine (BIBM)}, 
  title={Complementary-Disentangled Neural Generalization: a Robust Framework for Stable Brain-Computer Interfaces}, 
  year={2025},
  volume={},
  number={},
  pages={4230-4235},
  doi={10.1109/BIBM66473.2025.11356796}}

@inproceedings{gao2025cross,
  title={Cross-modal alignment between visual stimuli and neural responses in the visual cortex},
  author={Gao, Xing and Rong, Dazhong and He, Qinming},
  booktitle={Journal of Physics: Conference Series},
  volume={3147},
  number={1},
  pages={012026},
  year={2025},
  organization={IOP Publishing}
}

@inproceedings{rong2025improving,
  title={Improving Unsupervised Task-driven Models of Ventral Visual Stream via Relative Position Predictivity},
  author={Rong, Dazhong and Dong, Hao and Gao, Xing and Wei, Jiyu and Hong, Di and Hao, Yaoyao and He, Qinming and Wang, Yueming},
  booktitle={Proceedings of the Annual Meeting of the Cognitive Science Society},
  volume={47},
  year={2025}
}

@article{schneider2023learnable,
  title={Learnable latent embeddings for joint behavioural and neural analysis},
  author={Schneider, Steffen and Lee, Jin Hwa and Mathis, Mackenzie Weygandt},
  journal={Nature},
  volume={617},
  number={7960},
  pages={360--368},
  year={2023},
  publisher={Nature Publishing Group UK London}
}

@inproceedings{cascante2021curriculum,
  title={Curriculum labeling: Revisiting pseudo-labeling for semi-supervised learning},
  author={Cascante-Bonilla, Paola and Tan, Fuwen and Qi, Yanjun and Ordonez, Vicente},
  booktitle={Proceedings of the AAAI conference on artificial intelligence},
  volume={35},
  number={8},
  pages={6912--6920},
  year={2021}
}

@inproceedings{arazo2020pseudo,
  title={Pseudo-labeling and confirmation bias in deep semi-supervised learning},
  author={Arazo, Eric and Ortego, Diego and Albert, Paul and O’Connor, Noel E and McGuinness, Kevin},
  booktitle={2020 International joint conference on neural networks (IJCNN)},
  pages={1--8},
  year={2020},
  organization={IEEE}
}

@article{buttfield2006towards,
  title={Towards a robust BCI: error potentials and online learning},
  author={Buttfield, Anna and Ferrez, Pierre W and Millan, Jd R},
  journal={IEEE Transactions on Neural Systems and Rehabilitation Engineering},
  volume={14},
  number={2},
  pages={164--168},
  year={2006},
  publisher={IEEE}
}

@article{zhong2025deep,
  title={Deep joint subdomain alignment for unsupervised domain adaptation},
  author={Zhong, Zhenze and Wang, Dianyu and Zhou, Qiang and Lan, Ying},
  journal={Expert Systems with Applications},
  volume={262},
  pages={125602},
  year={2025},
  publisher={Elsevier}
}

@article{wei2021subdomain,
  title={Subdomain adaptation with manifolds discrepancy alignment},
  author={Wei, Pengfei and Ke, Yiping and Qu, Xinghua and Leong, Tze-Yun},
  journal={IEEE Transactions on Cybernetics},
  volume={52},
  number={11},
  pages={11698--11708},
  year={2021},
  publisher={IEEE}
}

@article{zhou2024mixstyle,
  title={Mixstyle neural networks for domain generalization and adaptation},
  author={Zhou, Kaiyang and Yang, Yongxin and Qiao, Yu and Xiang, Tao},
  journal={International Journal of Computer Vision},
  volume={132},
  number={3},
  pages={822--836},
  year={2024},
  publisher={Springer}
}

@article{ghifary2016scatter,
  title={Scatter component analysis: A unified framework for domain adaptation and domain generalization},
  author={Ghifary, Muhammad and Balduzzi, David and Kleijn, W Bastiaan and Zhang, Mengjie},
  journal={IEEE transactions on pattern analysis and machine intelligence},
  volume={39},
  number={7},
  pages={1414--1430},
  year={2016},
  publisher={IEEE}
}

\end{document}